\documentclass[runningheads]{llncs}

\usepackage{graphicx}
\usepackage{graphicx}
\usepackage{subfigure}
\usepackage{textcomp}
\usepackage{xcolor}
\usepackage{multirow}
\usepackage{enumitem}
\usepackage{graphicx}
\usepackage{subfigure}
\usepackage{enumerate}
\usepackage{pdflscape}
\usepackage{booktabs}
\usepackage{url}
\usepackage{xcolor}
\usepackage{color}

\usepackage{array}
\usepackage{mathtools}
\usepackage{makeidx}
\usepackage{times}
\usepackage{amssymb}
\usepackage{amsmath}
\usepackage{inputenc}

\usepackage{epstopdf}
\usepackage{mathrsfs}
\usepackage{etoolbox}
\usepackage{wrapfig,lipsum,booktabs}
\begin{document}

\title{Interpretable Visual Understanding with Cognitive Attention Network}

\author{Xuejiao Tang\inst{1}, Wenbin Zhang\inst{2}, Yi Yu\inst{3}, Kea Turner\inst{4}\\
Tyler Derr\inst{5}, Mengyu Wang\inst{6} and Eirini Ntoutsi\inst{7}}
\institute{$^1$Leibniz University of Hannover, Germany 
$^2$Carnegie Mellon University, USA\\
$^3$National Institute of Informatics, Japan
$^4$Moffitt Cancer Center, USA\\
$^5$Vanderbilt University, USA
$^6$Harvard Medical School, USA\\
$^7$Freie Universität Berlin, Germany\\
\email{\inst{1}xuejiao.tang@stud.uni-hannover.de, \inst{2}wenbinzhang@cmu.edu\\ \inst{3}yiyu@nii.ac.jp, \inst{4}Kea.Turner@moffitt.org, \inst{5}tyler.derr@vanderbilt.edu\\ 
\inst{6}mengyu\_wang@meei.harvard.edu, \inst{7}eirini.ntoutsi@fu-berlin.de}}

\authorrunning{X. Tang, W. Zhang, Y. Yu, K. Turner, T. Derr, M. Wang and E. Ntoutsi}

\maketitle              % typeset the header of the contribution
%
%\vspace{-0.5cm}
\begin{abstract}
\sloppy

While image understanding on recognition-level has achieved remarkable advancements, reliable visual scene understanding requires comprehensive image understanding on recognition-level but also cognition-level, which calls for exploiting the multi-source information as well as learning different levels of understanding and extensive commonsense knowledge. In this paper, we propose a novel Cognitive Attention Network (CAN) for visual commonsense reasoning to achieve interpretable visual understanding. Specifically, we first introduce an image-text fusion module to fuse information from images and text collectively. Second, a novel inference module is designed to encode commonsense among image, query and response. Extensive experiments on large-scale Visual Commonsense Reasoning (VCR) benchmark dataset demonstrate the effectiveness of our approach. The implementation is publicly available at \url{https://github.com/tanjatang/CAN}

\end{abstract}

\section{Introduction}
Visual understanding is an important research domain with a long history that attracts extensive models such as Mask RCNN~\cite{vuola2019mask}, ResNet~\cite{he2016deep} and UNet~\cite{barkau1996unet}. They have been successfully employed in a variety of visual understanding tasks such as action recognition, image classification, pose estimation and visual search~\cite{DBLP:journals/corr/PapandreouZKTTB17}. Most of them gain high-level understanding by identifying the objects in view based on visual input. However, reliable visual scene understanding requires not only recognition-level but also cognition-level visual understanding, and seamless integration of them. More specifically, it is desirable to identify the objects of interest to infer their actions, intents and mental states with an aim of having a comprehensive and reliable understanding of the visual input. While this is a natural task for humans, existing visual understanding systems suffer from a lack of ability for higher-order cognition inference~\cite{zellers2019recognition}.   

To improve the cognition-level visual understanding, recent research in visual understanding has shifted inference from recognition-level to cognition-level which contains more complex relationship inferences.

This directly leads to four major directions on cognition-level visual understanding research: 1) image generation~\cite{gregor2015draw}, which aims at generating images from given text description; 2) image caption~\cite{vinyals2015show}, which focuses on generating text description from given images; 3) visual question answering, which aims at predicting correct answers for given images and questions; 4) visual commonsense reasoning (VCR)~\cite{zellers2019recognition}, which additionally provides rational explanations along with question answering and has gained considerable attention~\cite{yu2020ernie}. Research on VCR typically necessitates pre-training on large scale data prior to performing VCR tasks. They usually fit well towards the properties that the pre-training data possessed but their generalization on other tasks are not guaranteed~\cite{chen2019uniter}. To remove the necessity of pre-training, another line of research focuses on directly learning the architecture of a system to find straightforward solutions for VCR~\cite{DBLP:journals/corr/abs-1910-14671}. However, these methods suffer commonsense information loss where the last hidden layer is taken as output while jointly encoding visual and text information.

In this paper, we focus on the generic problem of visual scene understanding, where the characteristics of multi-source information and different levels of understanding pose great challenges to comprehensive and reliable visual understanding: 1)~\textbf{Multi-source information.} Visual understanding entails information from different sources. It is difficult for the model to capture and fuse multi-source information and to infer the rationale based on the fusion of collective information and commonsense~\cite{natarajan2012multimodal}. 2)~\textbf{Various levels of understanding.} Cognition requires accumulation of an enormous reservoir of knowledge. Comprehensive cognition from limited datasets is even more challenging, and requires consideration of different levels of understanding~\cite{zellers2019recognition}. 3)~\textbf{Difficulty in learning commonsense.} The learning of commonsense from the dataset is a hard problem per se. Unlike humans who can learn an unlimited commonsense library from daily life effortlessly, learning extensive commonsense knowledge for a model is an open problem.

To address the above challenges, we propose a novel Cognitive Attention Network (CAN) for interpretable visual scene understanding. We first design a new multimodal fusion module to fuse image and text information based on guided attention. Then we introduce an co-attention network to encode the commonsense between text sequences and visual information, followed by an attention reduction module for redundant information filtering. The novelty of this research comes from four aspects:

%\vspace{-0.2cm}
\begin{itemize}
    \item A new VCR model for comprehensive and reliable visual scene understanding. 
    \item A new multimodal fusion method that jointly infers the multi-source information.
    \item A new co-attention network to encode commonsense.
    \item Extensive experiments comparing with state-of-the-art works and ablation studies.
\end{itemize}

The rest of the paper is organized as follows. Related studies are first discussed in Section~\ref{chap:relatedwork}. Section~\ref{nf} presents the notations and problem formulation. We describe our method in Section~\ref{chap:framework}, followed by the experimental results in Section~\ref{chap:experiment}. Finally, Section~\ref{chap:conclusion} concludes the paper.

% \vspace{-0.4cm}
\section{Related Work}
\label{chap:relatedwork}
% \vspace{-0.2cm}

From individual object level scene understanding~\cite{vuola2019mask} which aims at object instance segmentation and image recognition, to visual relationship detection~\cite{yang2019auto} which captures the relationship between any two objects in image or videos, state-of-the-art visual understanding models have achieved remarkable progress~\cite{DBLP:journals/corr/CarreiraZ17}. However, that is far from satisfactory for visual understanding as an ideal visual system necessitates the ability to understand the deep-level meaning behind a scene. Recent research on visual understanding has therefore shifted inference from recognition-level to cognition-level which contains more complex relationship inferences. Rowan et al.~\cite{zellers2019recognition} further formulated Visual Commonsense Reasoning as the VCR task, which is an important step towards reliable visual understanding, and benchmarked the VCR dataset. Specifically, the VCR dataset is sampled from a large sample of movie clips in which most of the scenes refer to logic inferences. For example, ``Why isn't Tom sitting next to David?'', which requires high-order inference ability about the scene to select the correct answer from available choices. Motivated studies generally fall into one of the following two categories based on the necessity of pre-training dataset.

The first line of research, pre-training approaches, trains the model on a large-scale dataset then fine-tunes the model for downstream tasks. The recent works include ERNIE-ViL-large~\cite{yu2020ernie} and UNITER-large~\cite{chen2019uniter}. While the former learns semantic relationship understanding for scene graph prediction, the latter is pre-trained to learn joint image-text representations. However, the generalizability of these models relies heavily on the pre-training dataset and therefore is not guaranteed.

Another line of research is independent of large-scale pre-training dataset, and instead studies the architecture of a system to find a straightforward solution for VCR. R2C~\cite{zellers2019recognition} is a representative example in this line of efforts in which attention based deep model is used for visual inferencing. More recently, a dynamic working memory based memory cells framework is proposed to provide prior knowledge for inference~\cite{tang2021}. Our model more closely resembles this method with two distinctions: i) a parallel structure is explicitly designed to relax the dependence on the previous cells, alleviating the drawback of information lose of long dependency memory cell for long sequences, and ii) a newly proposed co-attention network rather than dynamic working memory cell to ease model training but also to enhance the capability of capturing relationship between sentences and semantic information from surrounding words.

%\vspace{-0.2cm}
\section{Notations and Problem Formulation}
\begin{figure}
% \vspace{-0.8cm}
    \centering
    \includegraphics[width=11cm,height=7cm]{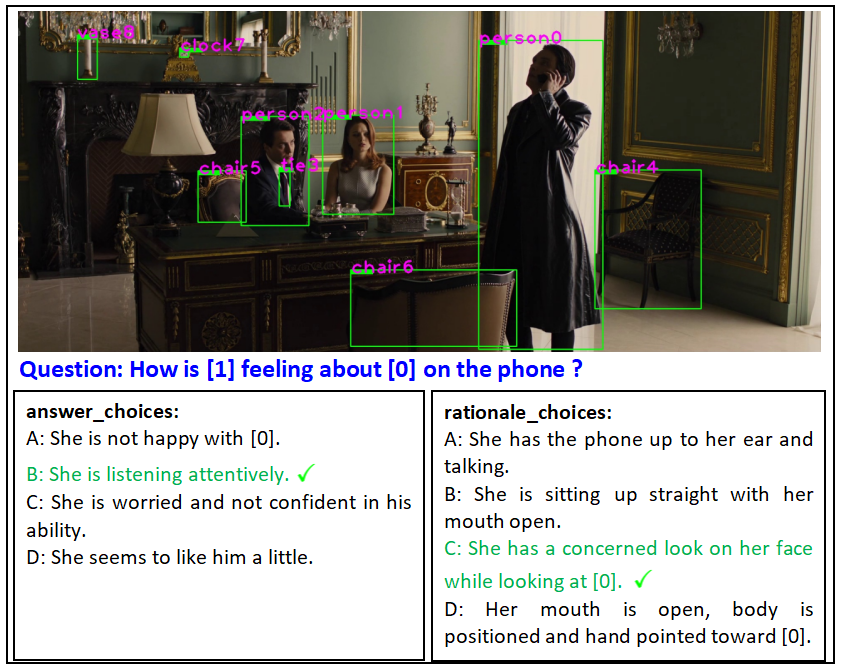}
    \caption{A VCR example with the correct answer and rationale highlighted in green.}
    \label{fig:VCR}
    \vspace{-0.5cm}
\end{figure}

% \vspace{-0.2cm}
\label{nf}
Given the input query $\textbf{q}:= \{q_1, q_2,..., q_m\}$ and the objects of the target image $\textbf{o}:= \{o_1, o_2, ..., o_n\}$, the general task of VCR is to sequentially predict one correct response from the responses represented as $\textbf{r}: = \{r_1, r_2,..., r_i\}$.
% and the reason the explains why answer is selected. 
Figure~\ref{fig:VCR} shows a typical VCR task, where \textbf{q} is to elicit information for Q (``How is [1] feeling about [0] on the phone?'') or both Q and its correct answer A (``She is listening attentively.'') depending on the specific sub-task discussed hereafter, \textbf{r} provides all possible answers or all reasons also depending on the specific sub-task, and \textbf{o} consists of objects of the image, i.e., person 0-2, tie 3, chair 4-6, clock 7 and vase 8. The three sub-tasks of VCR can then be represented as:

\begin{itemize}
    \item [1)] Q2A: is to predict the answer for the question. In this task, the inputs include: a) query \textbf{q}: question Q only, b) responses \textbf{r}: all possible answers, c) objects \textbf{o}, and d) given image, i.e., Figure~\ref{fig:VCR}. This sub-task needs to predict A based on the inputs.
    \item[2)] QA2R: is to reason why the answer is correct. Compared to the previous Q2A task, the query \textbf{q}, in addition to question Q, also includes the correct answer A and the responses \textbf{r} that are four given reasons. The aim of this sub-task is then to predict the correct reason R (``She has a concerned look on her face while looking at [0]'') for its input.
    \item[3)] Q2AR: is to integrate the results from the previous two tasks as the final result. The correct and wrong results will be shown and recorded for final performance evaluation. 
\end{itemize}

%\vspace{-0.3cm}
\section{The Proposed Framework}
\label{chap:framework}
%\vspace{-0.2cm}

\begin{figure}[!htbp]
    \centering
    \includegraphics[width=11cm]{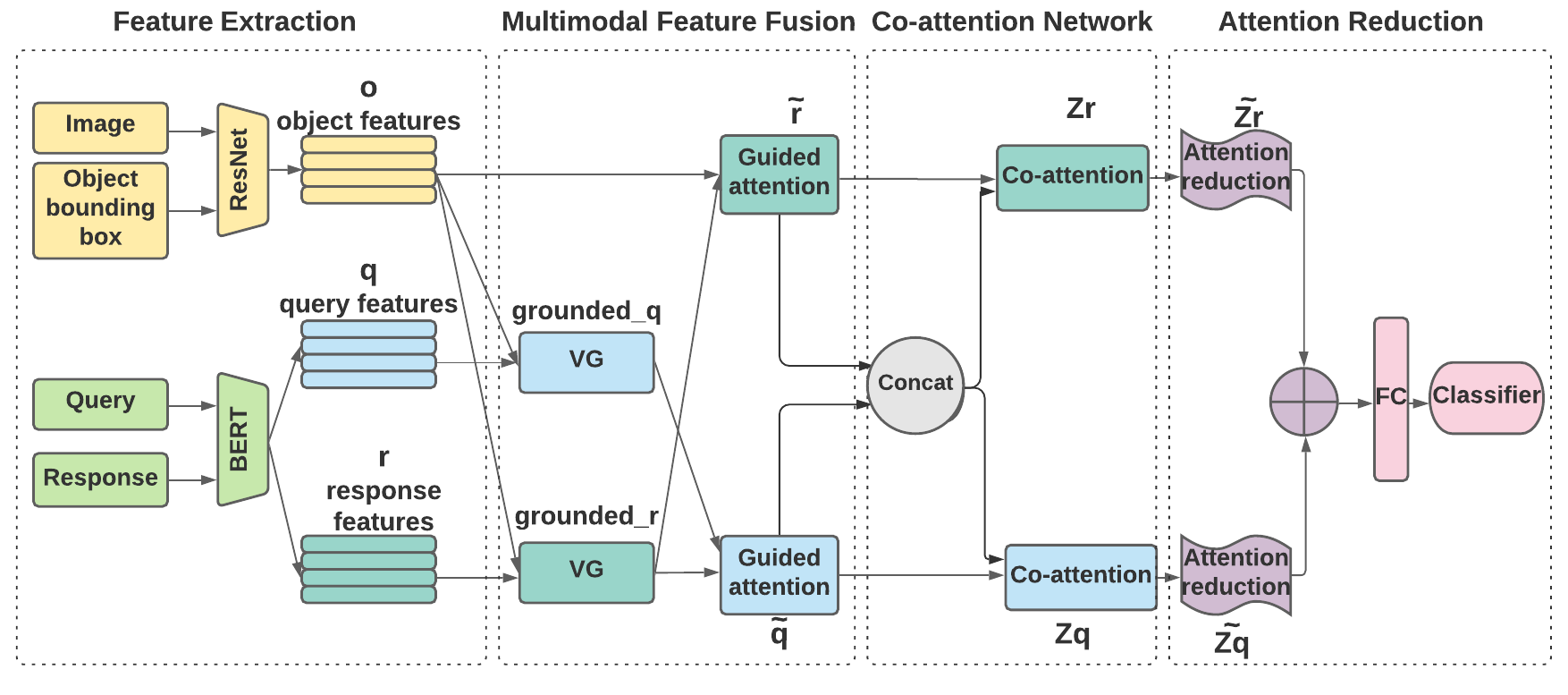}
    \vspace{-0.2cm}
    \caption{The architecture of the proposed CAN consists of four modules to achieve interpretable visual understanding.}
    \label{fig:framework}
    %\vspace{-0.5cm}
\end{figure}

The proposed Cognitive Attention Network (CAN) consists of four modules as shown in Figure~\ref{fig:framework}: a) \emph{feature extraction module} generates feature representations from the given multi-source image and text input, b) \emph{multimodal feature fusion module} integrates the extracted heterogeneous features; c) \emph{co-attention network} encodes the fused features; and d) \emph{attention reduction module} filters redundant information. The following subsections discuss the four modules in details. 

\vspace{-0.2cm}

\subsection{Feature Extraction}
\label{framework:obj-featurel}
Extracting informative features from multi-source information plays an important role in any machine learning application, especially in our context where the feature itself is one of the learning targets. As shown in Figure~\ref{fig:framework}, for the image feature extraction, the original image information source is the image along with its objects, which is given by means of related bounding boxes serving as a point of reference for objects within the images. The bounding boxes of given image and objects are then fed into the deep nets to obtain sufficient information from original image information source. Concretely, CAN extracts image features by a deep network backbone ResNet50~\cite{you2018imagenet} and fine-tunes the final block of the network after RoiAlign. In addition, the skip connection~\cite{he2016deep} is adopted to circumvent the gradient vanishing problem when training the deep nets.

In term of the text feature extraction, the original text information source includes Query (Q or Q together with A) and Response (given answers or reasons). The text information is then extracted in a dynamic way in which the attention mechanism is employed to encode information from words around them in parallel~\cite{devlin2018bert}, resulting text features including query features \textbf{q} and response features \textbf{r}.

% \vspace{-0.5cm}
\subsection{Multimodal Feature Fusion}
\label{framework:visual-grounding}

After features from heterogeneous information sources are extracted from the previous module, a multimodal feature fusion module is designed to fuse them, including: 1) a visual grounding unit to learn explicit information by aligning relevant objects with query and response; 2) a guided attention unit to learn implicit information that is omitted during visual grounding.

\textbf{Visual Grounding (VG).} To fuse the previously extracted heterogeneous features, i.e., related object features \textbf{o} along with text features \textbf{q} and \textbf{r}, a visual grounding module is designed to learn joint image-text representations explicitly.

To this end, VG firstly identifies related objects in query and response by using tags contained therein. Taking Figure~\ref{fig:VCR} as example, object features [person 0] and [person 1] are learned to match tags [0] and [1] in query \textbf{q} and responses \textbf{r}, while object features [person 2], [tie 3], [chair 4], [chair 5], [chair 6], [clock 7] and [vase 8] are omitted due to the lack of corresponding tags in \textbf{q} and \textbf{r}. Next, the aligned representations are fed into a one-layer bidirectional LSTM~\cite{huang2015bidirectional} to learn joint image-text representations. The learned image-query and image-response representations are denoted as $grounded\_q:=\{grounded\_q_1, grounded\_q_2, \cdots, grounded\_q_j\}$ and $grounded\_r:=\{grounded\_r_1, grounded\_r_2, \cdots, grounded\_r_j\}$, respectively.

\begin{figure}[!htbp]
    \centering
    % \vspace{-0.8cm}
    \includegraphics[width=\textwidth]{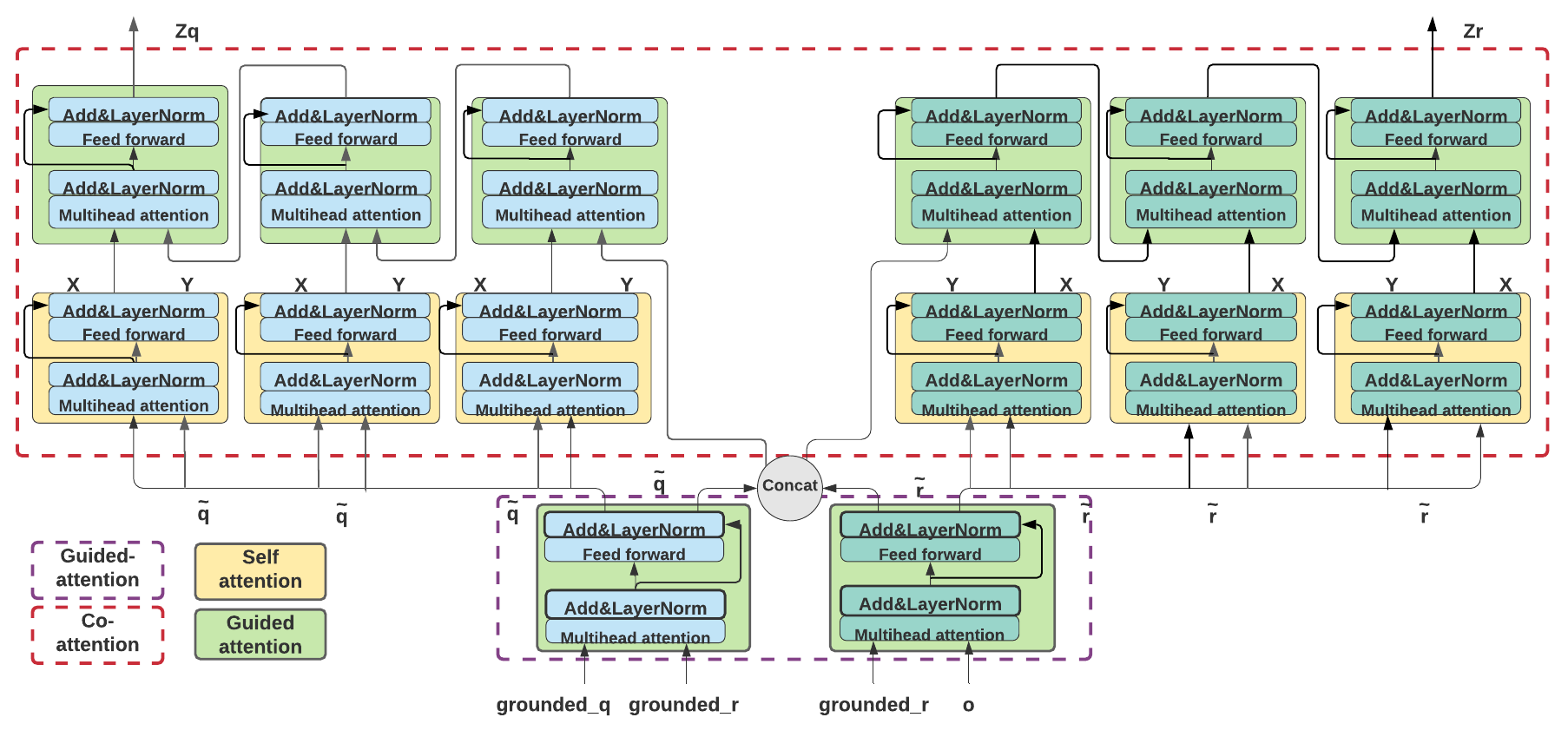}
    %\vspace{-0.2cm}
    \caption{Attention network of contextualizing feature representations. It consists of self-attention module and guided attention module to encode commonsense among image, query and response representations.}
    \label{fig:coattention}
    % \vspace{-0.8cm}
\end{figure}

\textbf{Guided Attention (GA).} After the VG stage, CAN learned an explicit joint image-text representations. However, the implicit information, which is important for commonsense inference including unidentified objects as well as reference relationship between grounded representations, is omitted. The guided attention module, shown as the two blocks within the purple dashed square in the bottom of Figure~\ref{fig:coattention}, is therefore designed to learn these implicit information, allowing for the attention on the two types of implicit but important correlations. Note that the unit of this guided attention module is also the atomic structure of the following co-attention network (c.f., Section~\ref{framework:co-attention module}). Specially, right hand side unit captures the implicit information between image-response representations $grounded\_r$ and image objects features $o$. Back to the running example in Figure~\ref{fig:VCR}, VG focuses on learning explicit information that is relevant to person 0 and person 1, and omits the explicit information associated with other objects, i.e., tie 3, chair 4-6, clock 7 and vase 8. This unit is designed to identify such implicit correlations between $grounded\_r$ and $o$. On the other hand, the left unit learns the implicit relationship between image-response representations $grounded\_r$ and image-query representations $grounded\_q$. For example in Figure~\ref{fig:VCR}, both ``[1]'' in the question (``How is [1] feeling about [0] on the phone'') and ``She'' in the answer (``She is listing attentively'') refer to identical person 1, but such implicit information is not learnable at VG stage. This unit accounts for such implicit correlations among $grounded\_r$ and $grounded\_q$. Note that attention can also be guided between $grounded\_q$ and $o$. However, $grounded\_q$ contains much lesser information than $grounded\_r$ as query normally entails lesser words and could be inferred from responses. Such an attention is therefore not considered to simplify the model with limited information loss. In the following, we will discuss the details of the proposed guided attention unit.

A guided attention unit is composed of a multi-head attention layer and a feed-forward layer. To speed up training, we additionally add LayerNorm for normalization behind both of these two layers. Recall that the aim of GA is to learn the omitted implicit information. To this end, GA first takes $\emph{o}$ and $\emph{grounded\_q}$ or $\emph{grounded\_r}$ as the input depending on the focused type of implicit information to guide the attention. Here, we employ the multi-head attention~\cite{vaswani2017attention} to guide this process. More specifically, multi-head attention consists of $h$ divided attention operations, referred as $heads$, through scaled dot-product attention. Formally put,   

\vspace{-0.4cm}
\begin{gather}
    \label{equ:multi-head3}
    MultiHead (Q_1, K_1, V_1) = Concat (head_1, ..., head_h)W^O 
\end{gather}

\noindent where $Q_1$ is $\emph{grounded\_r}$, both $K_1$ and $V_1$ are $\emph{o}$ or $\emph{grounded\_q}$, $W^{Q_1}_i, W^{K_1}_i, W^{V_1}_i, W^O$ are trainable linear transformation parameters, and $h$ is the total number of heads which can be formulated as:

\vspace{-0.4cm}
\begin{gather}
  \label{equ:multi-head2}
    head_i = Attention (Q_1W^{Q_1}_i, K_1W^{K_1}_i, V_1W^{V_1}_i)\\
    Attention (Q_1, K_1, V_1) = softmax(\frac{Q_1K_1^T}{\sqrt{d_k}})V_1
\end{gather}
% \vspace{-0.5cm}
% \begin{gather}
%   \label{equ:multi-head1}
%   Attention (Q_1, K_1, V_1) = softmax(\frac{Q_1K_1^T}{\sqrt{d_k}})V_1
% \end{gather}
\noindent where $T$ is the transpose operation, $d_k$ represents the dimension of input $K_1$, and $i$ is the $i$th head of total $h$ heads. In practise, $head_i$ outputs the attention weighted sum of the value vectors $V_1$ by softmax.

Next, the output of multi-head features are transformed by a feed-forward layer, which consists of two fully-connected layers with ReLU activation and dropout. Finally, GA outputs the fused multimodal representations $\widetilde{\textbf{q}}$ and $\widetilde{\textbf{r}}$ with weight information among \emph{o}, \emph{grounded\_q} and \emph{grounded\_r}.

\subsection{Co-attention Network}
\label{framework:co-attention module}

Given the fused image-text representations $\widetilde{\textbf{q}}$ and $\widetilde{\textbf{r}}$, we further propose a co-attention network to encode commonsense between the fused image-text representations for visual commonsense reasoning. The input of the network, in addition to $\widetilde{\textbf{q}}$ and $\widetilde{\textbf{r}}$, therefore further considers their joint representation $X$ defined as:

\vspace{-0.2cm}
\begin{equation}
\label{eq:concat qr}
    X = \widetilde{\textbf{q}} || \widetilde{\textbf{r}}
    %X = Concat (\widetilde{\textbf{q}}, \widetilde{\textbf{r}})
    %\vspace{-0.2cm}
\end{equation}

\noindent where $||$ is the concatenation operation. 

The red dashed square of Figure~\ref{fig:coattention} shows the structure of the co-attention network, consisting of two co-attention modules for attending query and response commonsense, respectively. In specific, the former is used for encoding commonsense between $X$ and $\widetilde{\textbf{q}}$, thus learning the attended commonsense for query jointly considers response. The latter then focuses on encoding commonsense between $X$ and $\widetilde{\textbf{r}}$, capturing the attended commonsense for response taking query into consideration. These two co-attention modules share the same structure, comprised of two sub-units: i) the self attention units, which are the blocks with yellow background in Figure~\ref{fig:coattention}, aiming at attending weighted information concerning each other within a sentence; ii) the blocks with green background depicted guided attention units to attend weighted information inter-sentence-wise as opposed to intra-sentence-wise attention of the self attention units.

\textbf{Self Attention.} The structure of self attention is similar to guided attention (c.f., Section~\ref{framework:visual-grounding}). The difference comes from self attention takes identical inputs, i.e., query $Q_1$, key $K_1$ and value $V_1$ are identical, for the sake of capturing pairwise relationship in a sequence. In details, pairwise relationship between samples in a sequence is learned by the multi-head attention layer. For input sequence $X = [x_1,x_2,...,x_m]$, the multi-head attention learns the relationship between $<x_i, x_j>$ and outputs attended representations. Subsequently, the attended representations are transformed by a feed-forward layer which contains two fully-connected layers with ReLU activation and dropout.

\textbf{Pairwise Guided Attention.} In comparison to self attention, pairwise guided attention focuses on inter-sentence-wise attention and can be regarded as guided attention learning weighted information among different sentences. When taking two different sentences representations $X = [x_1,x_2,...,x_m]$ and $Y = [y_1,y_2,...,y_m]$ as the inputs, $X$ is the query $Q_1$ while key $K_1$ and Value $V_1$ are $Y$, guiding the attention learning for $X$. Specifically, the multi-head layer in a guided attention unit attends the pairwise relationship between the two paired input sequences $<x_i, y_j>$ and outputs the attended representations. A feed-forward layer is then applied to transform the attended representations. The co-attention network finally outputs $Z_q$ and $Z_r$, which are attention information over both images and texts.

\vspace{-0.1cm}
\subsection{Attention Reduction}
\label{framework:Multi-Modal-Fusion}

After the previous multilayer data encoding, CAN now contains rich multi-source attention information. Among them, not all of them are necessarily to be innegligible. An attention reduction module is therefore further designed to select information with the most important attention weights. In details, the output of attention network $\underset{l \in \{q,r\}}{Z_l}$ are fed into a multilayer perceptron (MLP) to learn attention weights, outputting $\underset{l \in \{q,r\}}{\widetilde{Z_l}}$: 

\vspace{-0.2cm}
\begin{gather}
    \label{equ:wt}
    \widetilde{Z_l} = \sum_{i=1}^m \alpha_l^{i} z_{l}^i,~\alpha = softmax(MLP(Z_l))
\end{gather}

\noindent where $\alpha$ is the learned attention weights and $i$ is the position in a sequence.

For better gradient flow through the network, CAN also fuses the features by using LayerNorm on the sum of the final attended representations, 

\begin{equation}
    c =  LayerNorm (W_{x1}^T\widetilde{Z_q} + W_{x2}^T\widetilde{Z_r})
\end{equation}

\noindent where $W_{x1}^T$ and $W_{x2}^T$ are two trainable linear projection matrices. 

The fused feature $c$ is then projected by another FC layer for classification, which is used to find the correct answer and reason from given candidates, e.g., ``B. She is listening attentively'' and ``C. She has a concerned look on her face while looking at [0]'' among all other candidate answers and reasons in Figure~\ref{fig:VCR}.

\section{Experimental Results}
% \vspace{-0.2cm}
\label{chap:experiment}
This section evaluates the performance of our model in comparison to state-of-the-art visual understanding models. The experiments were conducted on a 64-bit machine with a 10-core processor (i9, 3.3GHz), 64GB memory with GTX 1080Ti GPU.

\subsection{Dataset}

The VCR dataset~\cite{zellers2019recognition} consists of 290k multiple-choice questions, 290k correct answers, 290k correct rationales and 110k images. The correct answers and rationales are labeled in the dataset with $>90\%$ of human agreements. As shown previously in Figure~\ref{fig:VCR}, each set consists of an image, a question, four available answer choices, and four reasoning choices. The correct answer and rationale are provided in the dataset as ground truth. 

\subsection{Understanding Visual Scenes}
We compare our method with several state-of-the-art visual scene understanding
models based on the mean average precision metric for the three Q2A, QA2R and Q2AR tasks, respectively, including: 1) MUTAN~\cite{DBLP:journals/corr/Ben-younesCCT17} proposes a multimodal based visual question answering approach, which parametrizes bi-linear interactions between visual and textual representations using Tucker decomposition; 2) BERT-base~\cite{vaswani2017attention} is a powerful pre-training based model in natural language field and is adapted for the commonsense reasoning; 3) R2C~\cite{zellers2019recognition} encodes commonsense between sentences with LSTM; 4) DMVCR~\cite{tang2021} trains a dynamic working memory to store the commonsense in training as well as using commonsense as prior knowledge for inference. Among them, BERT-base adopts pre-training method, while MUTAN, R2C and DMVCR are non pre-training methods. The obtained results are summarized in Table~\ref{experment results}.

\begin{table}[!htbp]
% \vspace{-0.5cm}
\centering
\renewcommand\arraystretch{1.0}
\begin{tabular}{|p{3.3cm}<{\centering}|p{2.8cm}<{\centering}|p{2.8cm}<{\centering}|p{2.8cm}<{\centering}|}
\hline
\textbf{Models} &\textbf{Q2A} &\textbf{QA2R} &\textbf{Q2AR} \\
\hline
\textbf{MUTAN~\cite{ben2017mutan}} & 44.4 & 32.0 & 14.6\\
\hline
\textbf{BERT-base~\cite{vaswani2017attention}} & 53.9 & 64.5 & 35\\
\hline
\textbf{R2C~\cite{zellers2019recognition}} & 61.9 & 62.8 &39.1\\
\hline
\textbf{DMVCR~\cite{tang2021}} & 62.4 & 67.5 & 42.3\\
\hline
\textbf{CAN} & \textbf{71.1} & \textbf{73.8} & \textbf{47.7}\\
\hline
\end{tabular}
\caption{Comparison of results between CAN and other methods on VCR dataset with the best performance marked in bold.}
\label{experment results}
\vspace{-0.9cm}
\end{table}

In these results, it is clear that CAN consistently outperforms other methods across all tasks and is the only method capable of handling all tasks properly. Specially, CAN outperforms MUTAN by a significant margin. This is expected as CAN incorporates a reasoning module in its encoder network to enhance commonsense understanding while MUTAN only focuses on visual question answering without reasoning. In addition, to alleviate the lost information when encoding long dependence structure for long sentences of other methods, CAN further encodes commonsense among sentences with
attention weights in parallel for a better information maintenance, which also leads to its superior performance over the others.

\subsection{Ablation Studies}

We also perform ablation studies to evaluate the performance of the proposed guided attention for multimodal fusion and co-attention network encoding. As one can see in Table~\ref{ablation results}, when taking out guided attention unit, the prediction result decreases 4.2\% in Q2A task and 5.7\% lower in QA2R task. It indicates guided attention can help the model learn implicit information from images, query and response representations, by attending the object in the images and the corresponding noun in the sentence. In addition, if we replace co-attention encoder network with LSTM encoder, the prediction result decreases 2.5\% in Q2A task and 4.6\% in QA2R task. Compared to LSTM keeping the memory among sentences, our proposed co-attention encoder network can attend the commonsense among various sentences and words with multi-head attention mechanism, thus capturing rich information from more aspects.   

\vspace{-0.3cm}
\begin{table}[]
% \vspace{-0.6cm}
\centering
\renewcommand\arraystretch{1.0}
\begin{tabular}{|p{4cm}<{\centering}|p{2.5cm}<{\centering}|p{2.5cm}<{\centering}|}
\hline
\textbf{Models} &\textbf{Q2A} &\textbf{QA2R} \\
\hline
\textbf{LSTM encoder} & 68.6 & 69.2 \\
\hline
\textbf{without GA} & 66.9 & 68.1 \\
% \hline
% \textbf{Prototype (Our method) } & 62.4 & \textbf{67.5} \\
\hline
\textbf{CAN} & \textbf{71.1} & \textbf{73.8} \\
\hline
\end{tabular}
\caption{Comparison of ablation studies.}
\label{ablation results}
\end{table}
\vspace{-0.9cm}

% \vspace{-0.5cm}
\subsection{Qualitative Results}
% \vspace{-0.1cm}
\label{exp:Qualitative results}

\begin{figure}[!htbp]
\label{fig:Qualitative results}

\addtocounter{subfigure}{0} 
\centering
\subfigure[Qualitative example 1. CAN predicts correct answer and rationale.]{
\label{fig:Qualitative result example 1}
\includegraphics[width = 9.35cm,height=6.7cm]{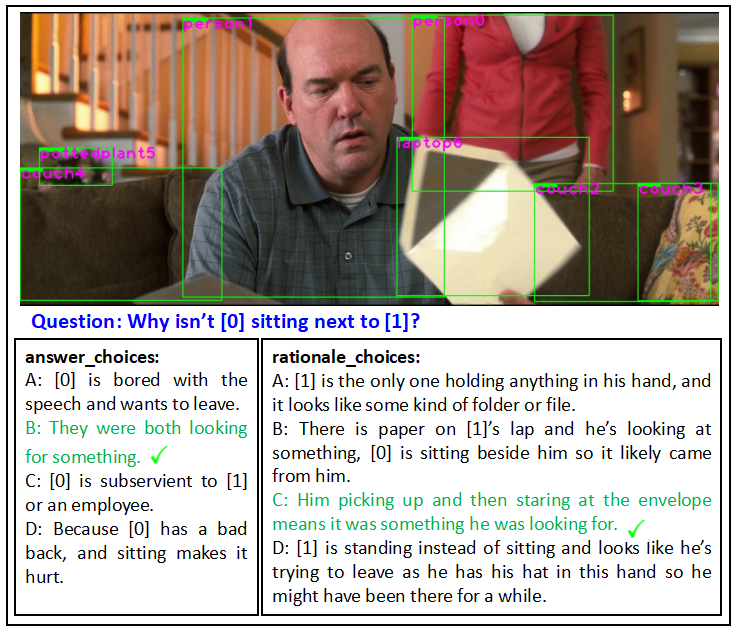}
}
\subfigure[Qualitative example 2. CAN predicts correct answer and rationale.]{
\label{fig:Qualitative result example 2}
    \includegraphics[width = 9.35cm,height=6.7cm]{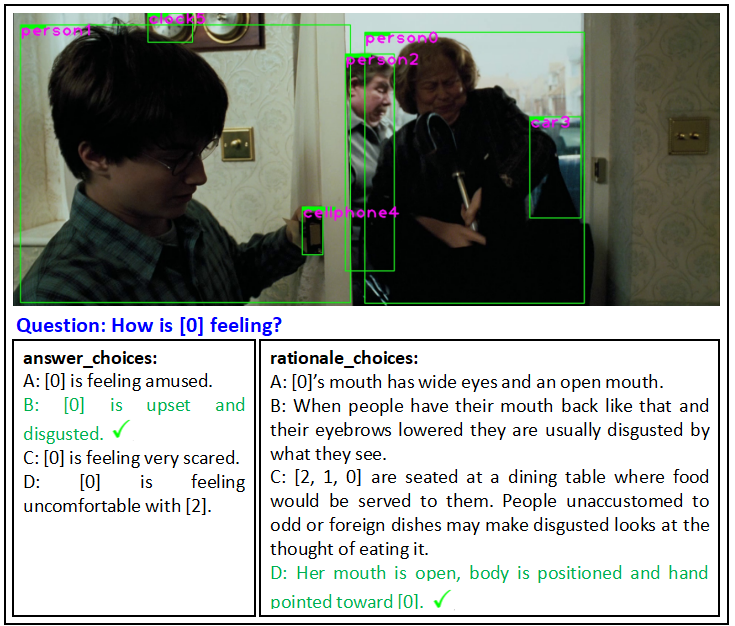}}

\subfigure[Qualitative example 3. CAN predicts incorrect answer but correct rational in Question 2.]{
\label{fig:Qualitative result example 3}
    \includegraphics[width = 9.35cm,height=6.7cm]{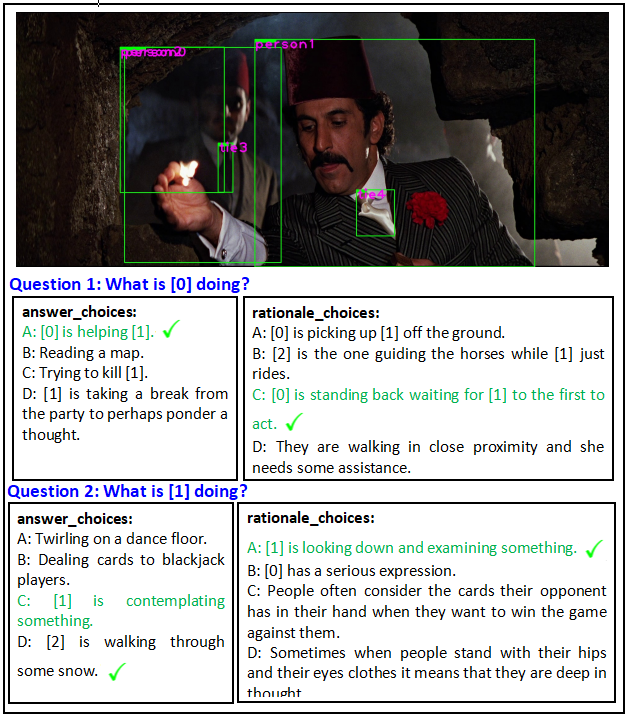}}
\caption{Qualitative examples. Prediction from CAN is marked by \textcolor{green}{\checkmark} while correct results are highlighted in green.}
\end{figure}

We evaluate the proposed framework with qualitative examples, which are shown in Figure~\ref{fig:Qualitative results}. The candidate with green color represents the correct choice along with the check mark \textcolor{green}{\checkmark} labeling the prediction by the proposed model. As the qualitative results show, our method works well for most of the visual scenes. For instance, in Figure~\ref{fig:Qualitative result example 1}, the query is ``Why isn't [person 0] sitting next to [person 1]?'', our model predicts the correct answer: ``B. They were both looking for something'', and the correct rationale ``C. Him picking up and then staring at the envelope means it was something he was looking for''. By co-attending the commonsense for [person 0] and [person 1] among the textual information in query, response and image representation, our model can select the correct answer and rationale for both Q2A and QA2R tasks.

Moreover, we can gain more insight into how the model understands the scene by co-attending the visual information and text information to predict the correct answer and rationale. For example in Figure~\ref{fig:Qualitative result example 2}, the question is ``How is [person 0] feeling?'', our model predicts the correct answer ``B. [person 0] is upset and disgusted'', and the correct rationale, ``D. Her mouth is open, body is positioned and hand pointed toward [person 0]''. This result shows that our model performs well by fusing multimodal features and co-attending the visual and textual information.

% \begin{wrapfigure}{c}{0.1\textwidth}\vspace{-0.3cm}
% 	\centering
% 	\addtocounter{subfigure}{0} 
% \centering
% \subfigure[Qualitative example 1. CAN predicts correct answer and rationale.]{
% \label{fig:Qualitative result example 1}
% \includegraphics[width = 9.35cm,height=6.7cm]{fig/QualitativeResult2.png}
% }
% \subfigure[Qualitative example 2. CAN predicts correct answer and rationale.]{
% \label{fig:Qualitative result example 2}
%     \includegraphics[width = 9.35cm,height=6.7cm]{fig/QualitativeResult1.png}}

% \subfigure[Qualitative example 3. CAN predicts incorrect answer but correct rational in Question 2.]{
% \label{fig:Qualitative result example 3}
%     \includegraphics[width = 9.35cm,height=6.7cm]{fig/QualitativeResult4.png}}
% \caption{Qualitative examples. Prediction from CAN is marked by \textcolor{green}{\checkmark} while correct results are highlighted in green.}
% \end{wrapfigure}

Figure~\ref{fig:Qualitative result example 3} shows two more challenging scenarios. CAN successfully predicted the correct answer and rationale for Question 1 but provided the incorrect answer with right rationale. Recall that question answering task (Q2A) and answer justification task (QA2R) are two separate tasks, and QA2R task performs on the condition that the correct answer is given. Therefore, the result of QA2R is independent of Q2A, and CAN can still predict the correct rationale in this challenging setting.

\section{Conclusion}
\label{chap:conclusion}
% \vspace{-0.2cm}
In this paper we propose a novel cognitive attention network for visual commonsense reasoning to achieve interpretable visual understanding. This work advances prior research by developing an image-text fusion module to fuse information between images and text as well as the design of a novel inference module to encode commonsense among image, query and response comprehensively. Extensive experiments on VCR benchmark dataset show the proposed method outperforms state-of-the-art by a wide margin. One promising future direction is to explore visual reasoning with fairness constraints~\cite{zhang2019faht}. 

\nocite{*}
\bibliographystyle{IEEEtran}
\bibliography{ref}
\end{document}